\documentclass[11pt]{article}
\usepackage{graphicx}
\usepackage{color}
\usepackage{amsmath}
\usepackage{amssymb}
\usepackage{amscd}
\usepackage{framed}
\usepackage{bbm}

\usepackage{fancyhdr,a4wide}
\usepackage{amsthm}

\newcommand{\R}{\mathbb{R}}

\newcommand{\E}{\mathbb{E}}
\newcommand{\M}{\mathbb{M}}
\newcommand{\Q}{\mathbb{Q}}
\newcommand{\B}{\mathbb{B}}

\newcommand{\eps}{\varepsilon}

\newtheorem{Theorem}{Theorem}[section]
\newtheorem{Lemma}[Theorem]{Lemma}

\newtheorem{Definition}[Theorem]{Definition}

\newtheorem{Corollary}[Theorem]{Corollary}
\newtheorem{Remark}[Theorem]{Remark}
\newtheorem{Example}[Theorem]{Example}

\numberwithin{equation}{section}

\def \proof {\noindent {\bf Proof.}\ \ }

\def \endproof
{{\mbox{}\nolinebreak\hfill\rule{2mm}{2mm}\par\medbreak}}
\def\IND{\mathbbm{1}}

\newcommand{\mP}{\mathcal{P}}

\def\IND{\mathbbm{1}}

\begin{document}
\title{Learning bounded subsets of $L_p$}
\author{
Shahar Mendelson \thanks{MSI, The Australian National University. Email: shahar.mendelson@anu.edu.au} \thanks{Part of this work was conducted while visiting the Faculty of Mathematics, University of Vienna, and the Erwin Schr\"{o}dinger Institute, Vienna.}
}

\maketitle

\begin{abstract}
We study learning problems in which the underlying class is a bounded subset of $L_p$ and the target $Y$ belongs to $L_p$. Previously, minimax sample complexity estimates were known under such boundedness assumptions only when $p=\infty$. We present a sharp sample complexity estimate that holds for any $p > 4$. It is based on a learning procedure that is suited for heavy-tailed problems.
\end{abstract}

\section{Introduction}
In the standard learning scenario one is given a class of functions $F$, defined on a probability space $(\Omega,\mu)$. Faced with an unknown random variable $Y$, the learner's goal is to find some function that approximates $Y$ almost as well as the best choice in $F$. More accurately, let $f^*$ be a best approximation of $Y$ in $F$ in the $L_2(\mu)$ sense; that is, if $X$ is distributed according to underlying measure $\mu$ then
$$
f^* \in {\rm argmin}_{f \in F} \E (f(X)-Y)^2,
$$
with the expectation taken with respect to the joint distribution of $X$ and $Y$. We assume throughout this note that such a minimizer exists, though it need not be unique.

In a more statistical language, $(f(X)-Y)^2$ is the price one pays for predicting $f(X)$ instead of $Y$, and $f^*$ is a function in $F$ that minimizes the ``average cost".
Hence, the learner is looking for some function $f \in L_2(\mu)$ such that
\begin{equation} \label{eq:excess-risk}
\E (f(X)-Y)^2 \leq \E (f^*(X)-Y)^2 + \eps = \inf_{h \in F} \E(h(X)-Y)^2 + \eps,
\end{equation}
where $\eps$ is the wanted accuracy.

What makes the problem of identifying such a function a difficult task is the limited information the learner has: not only is $Y$ unknown, but also the learner has no access to the underlying measure $\mu$ (i.e., to the distribution of $X$). Instead of having access to $X$ and $Y$, the learner is given a random sample, $(X_i,Y_i)_{i=1}^N$ selected independently according to the joint distribution $(X,Y)$. Using that sample, the learner has to produce a suitable function $f \in L_2(\mu)$. Naturally, finding a suitable function for every sample is unrealistic, and the best one can hope for is to be able to use most of the samples to generate a good approximation.

\begin{Definition} \label{def:learning}
Given a fixed sample size $N$, a learning procedure is a mapping $\Psi_N:(\Omega \times \R)^N \to L_2(\mu)$. Setting $\hat{f} = \Psi_N\left((X_i,Y_i)_{i=1}^N\right)$, the procedure performs with accuracy $\eps$ and confidence $1-\delta$ if, with probability at least $1-\delta$ over the samples $(X_i,Y_i)_{i=1}^N$, the conditional expectation satisfies that
$$
\E \left( \left(\hat{f}(X)-Y\right)^2 | (X_i,Y_i)_{i=1}^N \right) \leq \E (f^*(X)-Y)^2 +\eps.
$$
A learning procedure is \emph{proper} if $\hat{f} \in F$, and it is \emph{unrestricted} if it allowed to select functions outside $F$.
\end{Definition}

In what follows we associate a learning problem with a triplet $(F,X,Y)$ and one should keep in mind that out of the three only $F$ is known to the learner, whereas both $X$ and $Y$ are fixed but unknown. Also, at times we abuse notation by writing $f$ instead of $f(X)$, and by omitting the identity of the underlying probability space. For example, $\|f-h\|_{L_2}^2 = \E (f-h)^2(X)$, while $\|Y-f\|_{L_2}^2= \E (f(X)-Y)^2$. Finally, throughout this note absolute constants are denoted by $c, c_1,...$ etc. Their values may change from line to line. We denote $A \lesssim B$ if there is an absolute constant $c$ such that $A \leq c B$.

\vskip0.3cm

Learning problems of this flavour have been studied extensively over the last 50 years or so, and we refer the reader to the books \cite{DGL96,AnBa99,Kolt08} for a survey of their history. Nowadays it is well understood that the most obvious choice of a learning procedure, \emph{Empirical Risk Minimization} (ERM), in which one selects $\hat{f} \in F$ that best fits the data,
$$
\hat{f} = {\rm argmin}_{f \in F} \frac{1}{N}\sum_{i=1}^N \left(f(X_i)-Y_i\right)^2,
$$
is, in general, a poor choice.

There are two fundamental reasons behind the sub-optimality of ERM. Firstly, unless the triplet $(F,X,Y)$ satisfies some \emph{convexity condition}, the performance of any proper learning procedure will be far from optimal (see, e.g., the discussion in \cite{LugMen16,MenACM2}). As it happens, the convexity condition in question is satisfied in many interesting situations. Two such instances are when $F$ is convex and $(X,Y)$ is arbitrary (of course, under the assumption that $F \subset L_2(\mu)$); and when $X$ and $F \subset L_2(\mu)$ are arbitrary and $Y$ is independent additive noise, i.e., $Y=f_0(X)+W$ for some $f_0 \in F$ and $W \in L_2$ that is mean-zero and independent of $X$.

While the convexity condition may appear to be a rather minimal obstruction---many interesting examples do satisfy it---the second source of ERM's shortcoming is endemic: ERM fails even if the convexity condition is satisfied unless the random variables involved, namely $Y$ and $\{f(X) : f \in F\}$, are extremely light-tailed. In a heavier-tailed scenario, a typical sample $(X_i,Y_i)_{i=1}^N$ contains a nontrivial number of \emph{outliers}, and those outliers send ERM looking in the wrong part of $F$\footnote{Let us mention that a similar phenomenon occurs in \emph{mean estimation} problems, in which selecting the empirical mean of the random sample is a suboptimal guess of the true mean (see, for example, \cite{MR3909950} and the survey \cite{MR4017683} for more details).}.

The starting point of this note is a well-known result in Statistical Learning Theory, in which the situation seems highly favourable: the class $F$ is convex and all class members and the target $Y$ are bounded in $L_\infty$ by the same constant $M$. In particular, the two obstacles for using ERM are out of the way.

To formulate the result let us introduce the following notation: given a fixed triplet $(F,X,Y)$, the excess risk functional associated with a function $f$ is
$$
{\cal L}_f (X,Y) = (f(X)-Y)^2 - (f^*(X)-Y)^2,
$$
and, if $\hat{f}$ is the outcome of the learning procedure, set
$$
\E {\cal L}_{\hat{f}}= \E \left( \left(\hat{f}(X)-Y\right)^2 | (X_i,Y_i)_{i=1}^N \right) - \E (f^*(X)-Y)^2.
$$
\begin{framed}
Consider the triplet $(F,X,Y)$, where $F$ is convex and
$$
\sup_{f \in F} \|f\|_{L_\infty} \leq M, \ \ \ \ \|Y\|_{L_\infty} \leq M.
$$
Let
$$
N_0(r,\kappa) = \min \left\{ N \in \mathbbm{N} : \E \sup_{f \in F, \ \|f-f^*\|_{L_2} \leq  r} \left|\frac{1}{N}\sum_{i=1}^N \eps_i (f-f^*)(X_i) \right| \leq \kappa \frac{r^2}{M}\right\},
$$
where $(\eps_i)_{i=1}^N$ are independent, symmetric, $\{-1,1\}$-valued random variables that are independent of $(X_i)_{i=1}^N$, and the expectation is taken with respect to both $(X_i)_{i=1}^N$ and $(\eps_i)_{i=1}^N$.

\begin{Theorem} \label{thm:learning-L-infty} \cite{BaBoMe05,Kolt08}
There exist absolute constants $c_0$ and $c_1$ for which the following holds. Let $\eps$ and $\delta$ be the wanted accuracy and confidence parameters and let
$$
N \geq c_0 \left(N_0(\sqrt{\eps},c_1) + \frac{M^2}{\eps} \log\left(\frac{2}{\delta}\right)\right).
$$
Then ERM, upon receiving a sample $(X_i,Y_i)_{i=1}^N$ returns $\hat{f} \in F$ such that
$$
\E {\cal L}_{\hat{f}} \leq \eps \  {\rm with \ probability \ at \ least \ } 1-\delta.
$$
\end{Theorem}
\end{framed}
The question we focus on here is whether Theorem \ref{thm:learning-L-infty} can be extended in two significant directions: firstly, whether the assumption that both $F$ and $Y$ are bounded in $L_\infty$ can be relaxed to other $L_p$ spaces; and secondly, whether the convexity assumption can be removed completely. We show that the answer to both questions is positive, as long as $p > 4$.

\subsection{The difficulty of learning in $L_p$}
The proof of Theorem \ref{thm:learning-L-infty} relies heavily on the fact that all random variables involved are uniformly bounded in $L_\infty$. It allows one to use two crucial tools: Talagrand's concentration inequality for an empirical process indexed by a bounded subset of $L_\infty$ (\cite{MR1258865}, see also the book \cite{BoLuMa13}), and the contraction inequality for Bernoulli processes \cite{LeTa91}. Both are no longer valid when one departs from a bounded scenario and considers the significantly weaker assumption---that for some $p>4$,
$$
F \subset M B(L_p(\mu)) = \{f : \|f\|_{L_p} \leq M\} \ \ {\rm and} \ \ \|Y\|_{L_p} \leq M.
$$
Firstly, in such a situation one cannot hope to have sufficient concentration of empirical means around the true ones, and the analysis of the learning procedure has to be based on other methods. Secondly, and more significantly, the difficulty is not only technical, as the learning procedure itself has to be reconsidered: ERM has absolutely no chance of success under such weak assumptions---even if $F$ happens to be convex. The random variables involved can be heavy-tailed and the typical sample will contain many outliers, ``confusing" ERM.

Thanks to the recent progress in \cite{LugMen16,MenACM2} there are procedures, proper and unrestricted, that perform well in heavy-tailed situations. These procedures are based on some properties of the triplet, most notably that class members satisfy a uniform \emph{small-ball condition} or a uniform \emph{integrability condition} in the following sense:

\begin{Definition} \label{def:standard-assumption}
A function $f$ satisfies a small-ball condition with constants $\kappa>0$ and $0<\gamma<1$ if
\begin{equation} \label{eq:SB-def}
Pr( |f|(X) \geq \kappa \|f\|_{L_2}) \geq \gamma,
\end{equation}
It satisfies a $(\Gamma,\xi)$ integrability condition for $\Gamma>0$ and $0<\xi<1$ if
\begin{equation} \label{eq:tail-est-def}
\E f^2 \IND_{\{|f| \geq \Gamma \|f\|_{L_2}\}} \leq \xi \|f\|_{L_2}^2.
\end{equation}
A class $F \subset L_2(\mu)$ satisfies each one of the two conditions uniformly if \eqref{eq:SB-def} or \eqref{eq:tail-est-def} hold with the same constants for any $f=u-v$, where $u,v \in F \cup \{0\}$.
\end{Definition}
Intuitively, the role of the small-ball condition is to ensure that when $\|f-h\|_{L_2}$ is nontrivial, that fact is exhibited by the values $(|f-h|(X_i))_{i=1}^N$, allowing one to distinguish whether two functions are close or not based on the given sample.

Clearly, the integrability condition means that the contribution to $\|f\|_{L_2}$ of the set $\{ |f| \geq \Gamma \|f\|_{L_2}\}$ is not ``too big". It is straightforward to verify that a $(\Gamma,\xi)$ integrability condition is formally stronger than the small-ball condition (for the right constants).

Unfortunately, these two uniform conditions, which play a crucial role in the results of \cite{MenACM2}, do not hold even when $F$ consists of functions bounded almost surely by $M$, let alone when $F$ is merely a bounded subset of $L_p$ for some fixed $p > 4$.
\begin{Example} \label{eq:no-SB-L-infty}
Let $0<r < M$. Consider the function $f$ which satisfies that $Pr(f=M)=r^2/M^2$ and otherwise $f=0$. Then $\|f\|_{L_2}=r$ and $\|f\|_{L_p}=M$.

For any $0<\kappa < M/r = \|f\|_{L_\infty}/\|f\|_{L_2}$, \eqref{eq:SB-def} holds only for $\gamma=\|f\|_{L_2}^2/\|f\|_{L_\infty}^2$; whereas in \eqref{eq:tail-est-def} $\Gamma$ has to be at least $\|f\|_{L_\infty}/\|f\|_{L_2}$ for any $0<\xi<1$.
\end{Example}

Example \ref{eq:no-SB-L-infty} implies that the arguments used in \cite{MenACM2} to deal with classes of heavy-tailed functions fail in a bounded scenario: $F$ need not satisfy a nontrivial uniform small-ball condition or a nontrivial uniform integrability condition. Clearly, the situation is even worse when $F$ is a bounded subset of $L_p$ for some $4<p<\infty$.

\subsection{The main result}
As noted previously, once one departs from a light-tailed scenario, using ERM is out of the question. Therefore, the learning procedure that is used in what follows is closer in nature to the one from \cite{MenACM2}. At the same time, because triplets $(F,X,Y)$ need not satisfy \eqref{eq:SB-def} or \eqref{eq:tail-est-def}, the procedure from \cite{MenACM2} has to changed, as does some of the analysis.

Let us introduce some additional notation. Let $D$ be the unit ball in $L_2(\mu)$ and set $rD=\{f : \|f\|_{L_2} \leq r\}$. For a class $U$ and a fixed function $h$ set
$$
{\rm star}(U,h) = \left\{ \lambda u + (1-\lambda) h : 0 \leq \lambda \leq 1, \ u \in U\right\}.
$$
Thus, ${\rm star}(U,h)$ is the star-shaped hull of $U$ and $h$: the union of all intervals of the form $[u,h]$ for $u \in U$. Let
$$
U_{h,r} = {\rm star}(U-h,0) \cap rD = \{ w = \lambda(u-h) : u \in U, \ 0 \leq \lambda \leq 1, \ \|w\|_{L_2} \leq r\},
$$
and set
$$
\bar{U}=\frac{U+U}{2}=\left\{\frac{u+v}{2} : u, v \in U\right\}.
$$
Next, for a class $U$ and $h \in U$ let
\begin{equation} \label{eq:fixed-point}
r^*_\mathbbm{Q}(U,h,\kappa) = \inf \left\{ r>0 : \E \sup_{v \in U_{h,r}} \left|\frac{1}{N} \sum_{i=1}^N \eps_i v(X_i) \right| \leq \kappa r\right\}.
\end{equation}
In a similar fashion, for the triplet $(U,X,Y)$ let
\begin{equation} \label{eq:fixed-point-noise}
r^*_\mathbbm{M}(U,h,\kappa) = \inf \left\{ r>0 : \E \sup_{v \in U_{h,r}} \left|\frac{1}{N} \sum_{i=1}^N \eps_i \xi_iv(X_i) \right| \leq \kappa r^2\right\},
\end{equation}
where $u^* \in {\rm argmin}_{u \in U} \E (u(X)-Y)^2$ and $\xi_i=u^*(X_i)-Y_i$.

In both \eqref{eq:fixed-point} and \eqref{eq:fixed-point-noise}, the natural choice is $h=u^*$, and the fixed points  \eqref{eq:fixed-point} and \eqref{eq:fixed-point-noise} capture the ``critical level" associated with a given triplet, around $u^*$. The meaning of that critical level is simple: by the star-shape property of ${\rm star}(U,h)$, the sets $U_{h,r}$ become ``richer" the smaller $r$ is. As a result, the fixed point $r_{\mathbbm{Q}}$ and is the smallest radius $r$ for which the oscillation of the random process in question is not too big: for example, if $r \geq  r^*_\mathbbm{Q}(U,h,\kappa)$, the oscillation of $U_{h,r}$ in \eqref{eq:fixed-point} is at most $\kappa r$, while if $r < r^*_\mathbbm{Q}(U,h,\kappa)$ the reverse inequality holds. A similar phenomenon holds for the oscillation in \eqref{eq:fixed-point-noise}. Therefore, when $r \geq \max\{r^*_\mathbbm{Q},r^*_\mathbbm{M}\}$ one has sufficient control on the two oscillations, and that is used to control the quadratic (corresponding to $\mathbbm{Q}$) and multiplier (corresponding to $\mathbbm{M}$) components of the excess loss functional (see the discussion in \cite{MenACM1,LugMen16,MenACM2} for more details).

Here, just as in Theorem \ref{thm:learning-L-infty}, it is more convenient to use the notion of sample complexity: given wanted accuracy and confidence parameters $\eps$ and $\delta$, the sample complexity is the smallest sample size that is required to ensure that the procedure performs with the wanted accuracy and confidence.

Let us define the ``sample complexity" versions of \eqref{eq:fixed-point} and \eqref{eq:fixed-point-noise}: for a triplet $(U,X,Y)$, set

\begin{equation} \label{eq:N-int}
N_{\mathbbm{Q}}(U,r,\kappa) = \min\left\{ N \in \mathbbm{N}: \E \sup_{v \in U_{u^*,r}} \left|\frac{1}{N} \sum_{i=1}^N \eps_i v(X_i) \right| \leq \kappa r\right\},
\end{equation}
and
\begin{equation} \label{eq:N-ext}
N_{\mathbbm{M}}(U,r,\kappa) = \min \left\{ N \in \mathbbm{N}: \E \sup_{v \in U_{u^*,r}} \left|\frac{1}{N} \sum_{i=1}^N \eps_i \xi_iv(X_i) \right| \leq \kappa r^2 \right\}.
\end{equation}
Again, thanks to the star-shape property, it is straightforward to verify that if
$$
N \geq 2 \max\{N_{\mathbbm{Q}}(U,r,\kappa_1),N_{\mathbbm{M}}(U,r,\kappa_2)\}
$$
then
$$
\E \sup_{v \in U_{u^*,r}} \left|\frac{1}{N} \sum_{i=1}^N \eps_i v(X_i) \right| \leq \kappa_1 r \ \ {\rm and} \ \ \E \sup_{v \in U_{u^*,r}} \left|\frac{1}{N} \sum_{i=1}^N \eps_i \xi_iv(X_i) \right| \leq \kappa_2 r^2.
$$

To put these conditions in some perspective, let us reformulate Theorem \ref{thm:learning-L-infty} using $N_{\mathbbm{Q}}$ and $N_{\mathbbm{M}}$:
\begin{Theorem} \label{thm:L-infty-reform}
There are absolute constants $c_1$ and $c_2$ for which the following holds. Let $F \subset MB(L_\infty)$ be a convex class and assume that  $\|Y\|_{L_\infty} \leq M$. Given $\eps>0$ and $0<\delta<1$, set $\kappa_1=c_1\sqrt{\eps}/M$ and $\kappa_2=c_2$. If
\begin{equation} \label{eq:reform-L-infty-bound}
N  \geq 2\max\{N_{\mathbbm{Q}}(F,\sqrt{\eps},\kappa_1),N_{\mathbbm{M}}(F,\sqrt{\eps},\kappa_2)\} + c_3 \frac{M^2}{\eps}\log\left(\frac{2}{\delta}\right)
\end{equation}
then ERM returns $\hat{f}$ which satisfies that
$$
\E {\cal L}_{\hat{f}} \leq \eps \ {\rm with \ probability \ at \ least \ } \  1-\delta.
$$
\end{Theorem}
The reason that this is indeed a reformulation of Theorem \ref{thm:learning-L-infty} is one of the key features of an empirical process indexed by a bounded subset of $L_\infty$: the contraction inequality for Bernoulli processes, which implies that since $\|f^*(X)-Y\|_{L_\infty} \leq 2M$, one has that
$$
\E \sup_{v \in F_{f^*,r}} \left|\frac{1}{N} \sum_{i=1}^N \eps_i \xi_iv(X_i) \right| \leq 2M \E \sup_{v \in F_{f^*,r}} \left|\frac{1}{N} \sum_{i=1}^N \eps_i v(X_i) \right|.
$$
As a result, $N_{\mathbbm{Q}}(F,\sqrt{\eps},\kappa_1)$ dominates $N_{\mathbbm{M}}(U,\sqrt{\eps},\kappa_2)$ and the latter can be omitted from the sample complexity estimate.

\begin{Remark}
It is important to stress that while Theorem \ref{thm:learning-L-infty} is optimal in the minimax sense, it does not mean that its outcome is optimal for any triplet $(F,X,Y)$ that belongs to the bounded framework. Rather, minimax optimality means that there are cases in which \eqref{eq:reform-L-infty-bound} matches (up to a multiplicative absolute constant) a lower bound. It does not rule out that a smaller sample size could suffice under some additional assumptions on the triplet.
\end{Remark}

\vskip0.3cm

Before formulating the main result, let us see what one can hope for, taking into account that the estimate in Theorem \ref{thm:learning-L-infty} is minimax optimal.
\begin{Example}
Let $f:\Omega \to \{0,K\}$ which satisfies $Pr(f=K)=r^2/K^2$ for the choice $K = M^{p/(p-2)}/r^{2/(p-2)}$. Then $\|f\|_{L_2}=r$ and $\|f\|_{L_p} =M$. Thus, there are classes of functions $F \subset M(B(L_p(\mu)))$ which also satisfy that for every $r>0$
$$
F \cap r S(L_2) \subset \left\{ f: \|f\|_{L_\infty} = \frac{M^{{p}/{(p-2)}}}{r^{{2}/{(p-2)}}}\right\}.
$$
 Based on Theorem \ref{thm:learning-L-infty} and using the (level-dependent) $L_\infty$ estimate of $K(r) = \frac{M^{{p}/{(p-2)}}}{r^{{2}/{(p-2)}}}$
the fixed point condition in Theorem \ref{thm:learning-L-infty} becomes
\begin{equation} \label{eq:L-p-to-bounded-SC1}
\E \sup_{v \in F_{f^*,\sqrt{\eps}}} \left|\frac{1}{N} \sum_{i=1}^N \eps_i v(X_i) \right| \leq  \frac{\eps} {K(\sqrt{\eps})} = \sqrt{\eps} \cdot \left(\frac{\sqrt{\eps}}{M}\right)^{\frac{p}{p-2}},
\end{equation}
and the corresponding sample complexity term is
$$
N_{\mathbbm{Q}} \left(F,\sqrt{\eps},\left(\frac{\sqrt{\eps}}{M}\right)^{\frac{p}{p-2}}\right).
$$
Moreover, the term in the sample complexity estimate that depends on the confidence parameter $\delta$ becomes
\begin{equation} \label{eq:L-p-to-bounded-SC2}
\frac{K^2(\sqrt{\eps})}{\eps} \log \left(\frac{2}{\delta}\right) = \left(\frac{M^2}{\eps}\right)^{\frac{p}{p-2}} \log\left(\frac{2}{\delta}\right).
\end{equation}
\end{Example}

Equations \eqref{eq:L-p-to-bounded-SC1} and \eqref{eq:L-p-to-bounded-SC2} represent the best  one could hope for when considering only very special classes that are bounded in $L_p$, and which have ``level sets" that are well bounded in $L_\infty$.  It is somewhat surprising that these conditions, together with the fixed point condition \eqref{eq:N-ext} actually suffice for addressing \emph{any} learning problem associated with a triplet $(F,X,Y)$, consisting of $F \subset MB(L_p(\mu))$ and $\|Y\|_{L_p} \leq M$, and without assuming that $F$ needs to be convex.


\begin{Theorem} \label{thm:main}
There exist absolute constants $c_0$, $c_1$, $c_2$ and $c_3$ for which the following holds. Let $p > 4$, and consider a triplet $(F,X,Y)$ such that $F \subset M B(L_p(\mu))$ and $\|Y\|_{L_p} \leq M$. There is a procedure $\hat{f}$ that receives as data the wanted accuracy and confidence parameters $\eps$ and $\delta$, the values $M$ and $p$, and a sample $(X_i,Y_i)_{i=1}^N$.
\begin{description}
 \item{$\bullet$} If $F$ is convex then $\hat{f}$ is proper. Let
 \begin{equation*}
N_0 = c_0 \left( 2\max\left\{N_{\mathbbm{Q}}\left(F,\sqrt{\eps},c_1\left(\frac{\sqrt{\eps}}{M}\right)^{\frac{p}{p-2}}\right) ,N_{\mathbbm{M}}(F,\sqrt{\eps},c_2)\right\}
+\left(\frac{M^2}{\eps}\right)^{\frac{p}{p-2}}\log\left(\frac{2}{\delta}\right) \right),
\end{equation*}
and if $N \geq N_0$ then
$$
\E {\cal L}_{\hat{f}} \leq c_3\eps \ {\rm with \ probability \ at \ least \ }  1 -\delta.
$$
\item{$\bullet$} For a general class $F$, $\hat{f}$ is unrestricted. Let ${\cal U}$ be the collection of all subsets of $\bar{F}=(F+F)/2$ that contain $f^*$. Let
\begin{equation*}
N_0 = c_0\left(\sup_{U \in {\cal U}} \max\left\{N_{\mathbbm{Q}}\left(U,\sqrt{\eps},c_1\left(\frac{\sqrt{\eps}}{M}\right)^{\frac{p}{p-2}}\right) ,N_{\mathbbm{M}}(U,\sqrt{\eps},c_2)\right\}+
\left(\frac{M^2}{\eps}\right)^{\frac{p}{p-2}}\log\left(\frac{2}{\delta}\right)\right),
\end{equation*}
where the supermum is taken oven all triplets $\{ (U,X,Y) : U \in {\cal U}\}$. If $N \geq N_0$ then
$$
\E {\cal L}_{\hat{f}} \leq c_3\eps  \ {\rm with \ probability \ at \ least \ } 1 -\delta.
$$
\end{description}
\end{Theorem}

\begin{Remark}
A version of Theorem \ref{thm:main} holds for any $p>2$, but the true situation when $p \in (2,4]$ is far from clear. In that range one requires an additional restrictive assumption and a slightly different procedure---see Remark \ref{rem:p-2-4} for more information.

The question of the optimal estimate in the range $(2,4]$ is left for future study.
\end{Remark}

The first part of Theorem \ref{thm:main} is a direct extension of Theorem \ref{thm:learning-L-infty}, as can be seen by taking $p \to \infty$. At the same time, the procedure is totally different: ERM has no chance of success in this  heavy-tailed situation.

The second part of Theorem \ref{thm:main} looks strange at a first glance, but its nature will become clear once the procedure is described in more detail (see Section \ref{sec:the-procedure}, below). Roughly put, the procedure has two stages: first, it uses the first half of the given sample to select a (data-dependent) subset of $F_1 \subset F$, which has some useful features---among which, it contains $f^*$. Then, the same procedure is performed using the second half of the sample in $\bar{F}_1=(F_1+F_1)/2$, resulting in $F_2 \subset \bar{F}_1$; $\hat{f}$ is chosen as any element in $F_2$. One can show that the success of the procedure is guaranteed with the wanted accuracy and confidence once the sample is large enough to ``deal with" the triplet $(F,X,Y)$ and then with the triplet $(\bar{F}_1,X,Y)$. However, $\bar{F}_1$ is, in itself, random, and there is no way in which one may pre-determine its identity. The one known feature of $F_1$ is that it contains $f^*$, and thus, a natural way of ensuring that the given sample is large enough is by taking the largest sample size that is needed for the procedure to ``deal with" an arbitrary subset of $\bar{F}$ that contains $f^*$.

\section{The procedure} \label{sec:the-procedure}
The procedure is based on a modification of the one introduced in \cite{MenACM2} with some significant modifications. It receives as input the wanted accuracy and confidence parameters $\eps$ and $\delta$; the value of $p$; the uniform bound on the $L_p$ norms of the random variables involved which is denoted by $M$; and a sample $(X_i,Y_i)_{i=1}^{4N}$. It also receives tuning parameters $\theta_1,...,\theta_4$ that have to be specified in advance, and turn out to be well-chosen absolute constants.

The sample is split to three sub-samples of cardinality $N$, and set
\begin{equation} \label{eq:condition-on-n-m-1}
m = \theta_1 \left(\frac{M^2}{\eps}\right)^{\frac{p}{p-2}} \ \ {\rm and} \ \ n = \frac{N}{m},
\end{equation}
where without loss of generality one assumes that both $m$ and $n$ are integers. Each sub-sample of cardinality $N$ is split to blocks $I_1,...,I_n$, each one of cardinality $m$.

The reason behind this choice will be made clear in what follows. Note that if
\begin{equation} \label{eq:sample-complexity-1}
N \gtrsim  \left(\frac{M^2}{\eps}\right)^{\frac{p}{p-2}} \log\left(\frac{2}{\delta}\right)
\end{equation}
then $n \gtrsim \log(2/\delta)$, and indeed, the probability estimate we establish will be $1-\exp(-cn)$ for an absolute constant $c$.


\vskip0.3cm

The first component in the procedure, denoted by $\mP_1$, is a \emph{distance oracle}, whose goal is to give ``educated guesses" of distances $\|f-h\|_{L_2}$ between any two class members. Formally, given $(X_i)_{i=1}^N$, for any $f,h \in \bar{F}=(F+F)/2$ and $1 \leq j \leq n$ put
$$
\mu_j(f,h) = \left|\frac{1}{\sqrt{m}} \sum_{i \in I_j} \eps_i (f-h)(X_i) \right|,
$$
where $(\eps_i)_{i=1}^N$ are independent, symmetric, $\{-1,1\}$-valued random variables that are independent of $(X_i)_{i=1}^{N}$. Set
$$
\mP_1(f,h) = {\rm Med}\left(\mu_1(f,h),...,\mu_n(f,h)\right)
$$
to be a median of the values $(\mu_j(f,h))_{j=1}^n$.

\vskip0.3cm
The second component of the procedure, $\mP_2$, receives as data the output of $\mP_1$, an independent sample of cardinality $N$ and $H \subset \bar{F}$. Consider the triplet $(H,X,Y)$ and for $h,f \in H$,  and $1 \leq j \leq n$ let
$$
\B_{h,f}(j)=\frac{1}{m}\sum_{i \in I_j} (h(X_i)-Y_i)^2 - \frac{1}{m}\sum_{i \in I_j} (f(X_i)-Y_i)^2.
$$

\begin{Definition} \label{def:win}
Set $r^2=\eps$ to be the wanted accuracy. Define $f \succ h$ if, for more than $n/2$ of the  blocks $I_j$, one has
\begin{equation}
\begin{cases}
\B_{h,f}(j) \geq -\theta_2 r^2 & \mbox{when} \ \ \ \mP_1(h,f) \leq \theta_4 r, \ \ \ {\rm or}
\\
\B_{h,f}(j) \geq -\theta_3 \mP_1^2(h,f) & \mbox{when} \ \ \ \mP_1(h,f)>\theta_4 r.
\end{cases}
\end{equation}
Let
\begin{equation} \label{eq:P-2-def}
\mP_2(H)=\left\{f \in H : \ f \succ h \ {\rm for \ every } \ h \in H\right\}.
\end{equation}
\end{Definition}

With the two components in place, the learning procedure is:
\begin{framed}
\begin{description}
\item{$(1)$} Use the first part of the sample, $(X_i)_{i=1}^N$ to obtain the set of values $\{ \mP_1(h,f) : h,f \in F \}$.
\item{$(2)$} With the values $\{ \mP_1(h,f) : h,f \in F \}$, the sample $(X_i,Y_i)_{i=N+1}^{2N}$ and $F$ as input, use $\mP_2$ to generate the class $F_1=\mP_2(F)$.
\item{$(3)$} Repeat $(1)$ and $(2)$ for $\bar{F_1}=(F_1+F_1)/2$: use the two halves of $(X_i,Y_i)_{i=2N+1}^{4N}$ to first generate $\{ \mP_1(h,f) : h,f \in \bar{F}_1 \}$ and then $F_2=\mP_2(\bar{F}_1)$.
\item{$(4)$} Select $\hat{f}$ to be any function in $F_2$.
\end{description}
The claim of Theorem \ref{thm:main} is that if $N \geq N_0$ one has that
$$
\E{\cal L}_{\hat{f}} \lesssim \eps \ \ \ {\rm with \ probability \ at \ least \ } 1-\delta.
$$
\end{framed}

\begin{Remark}
The main novelty in the procedure is the distance oracle $\mP_1$, which is different from the one used in \cite{MenACM2}. The distance oracle from \cite{MenACM2} need not perform well when all that is known is that $F$ is a bounded subset of $L_p(\mu)$.
\end{Remark}

Most of the groundwork necessary for proving Theorem \ref{thm:main} was carried out in \cite{MenACM2,MenStudia}. Most notably, it was shown in \cite{MenACM2} that ensuring this type of learning procedure performs well can be done by verifying certain features of natural random processes indexed by subsets of $\bar{F}$---as we now describe.

%

\subsection{Properties of $\mP_1$}
\begin{Definition} \label{def:P-1-distace oracle}
A triplet $(H,X,Y)$ satisfies $(\clubsuit)$ with constants $\alpha<1$, $\beta>1$, $0<\delta<1$ and $r>0$ if the following holds. Setting $h^* \in {\rm argmin}_{h \in H} \E (h(X)-Y)^2$, there is an event ${\mathcal A}^\prime$ of probability at least $1-\delta/4$ on which, for any $h \in H$ one has
\begin{description}
\item{$\bullet$} If $\mP_1(h,h^*) \geq \beta r$ then  $\beta^{-1}\mP_1(h,h^*) \leq \|h-h^*\|_{L_2} \leq \alpha^{-1}\mP_1(h,h^*)$.
\item{$\bullet$}  If $\mP_1(h,h^*) < \beta r$ then $\|h-h^*\|_{L_2} \leq (\beta/\alpha)r$.
\end{description}
\end{Definition}
Property $(\clubsuit)$ implies that on a large event, ${\mathcal A}^\prime$, if $\mP_1(h,h^*)$ is large enough then it is a two-sided isomorphic estimate of $\|h-h^*\|_{L_2}$; and otherwise, $h$ and $h^*$ are relatively close.

\subsection{Properties of $\mP_2$}
The quadratic and multiplier components of the squared excess loss play a key role in the study of $\mP_2$. Note that for a partition $I_1,...,I_n$ of $\{1,...,N\}$ one has that
$$
\sum_{i \in I_J} (h(X_i)-Y_i)^2 - (h^*(X_i)-Y_i)^2 = \sum_{i \in I} (h-h^*)^2(X_i) + 2\sum_{i \in I_j} (h^*(X_i)-Y_i)(h-h^*)(X_i),
$$
and set
$$
\Q_{h,h^*}(j) = \frac{1}{m} \sum_{i \in I} (h-h^*)^2(X_i) \ \ {\rm and} \ \ \M_{h,h^*}(j) = \frac{2}{m}\sum_{i \in I_j} (h^*(X_i)-Y_i)(h-h^*)(X_i).
$$

\begin{Definition} \label{def:P-2-comparing-means}
The triplet $(H,X,Y)$ satisfies $(\diamondsuit)$, $(\heartsuit)$ and $(\spadesuit)$ with constants $\gamma>0$, $0<\nu<1$, $r>0$ and $0<\delta<1$, if the following holds. There is an event ${\mathcal A}^{\prime \prime}$ of probability at least $1-\delta/4$, on which, for every $h \in H$,
\begin{description}
\item{$\bullet$} If $\|h-h^*\|_{L_2} \geq r$ then on more than $n/2$ of the blocks $I_j$,
\begin{align*}
(\diamondsuit) \ \ \ &  \Q_{h,h^*}(j) \geq (1-\nu)\|h-h^*\|_{L_2}^2 \ \ {\rm and}
\\
(\heartsuit) \ \ \ & \M_{h,h^*} - \E \M_{h,h^*} \geq -\nu \|h-h^*\|_{L_2}^2.
\end{align*}

\item{$\bullet$} If $\|h-h^*\|_{L_2} \leq (\beta/\alpha)r$ then on more than $n/2$ of the blocks
$$
(\spadesuit) \ \ \ |\M_{h,h^*}(j)-\E \M_{h,h^*} | \leq \gamma r^2.
$$
\end{description}
\end{Definition}
Finally,
\begin{Definition} \label{def:suitable}
A  triplet $(F,X,Y)$ is called \emph{suitable} with constants $\alpha$,$\beta$,$\gamma$, $\delta$, $\nu$ and $r$ if for every subset $H \subset \bar{F}$ that contains $f^*$, the triplet $(H,X,Y)$ satisfies $(\clubsuit)$ with constants $\alpha$, $\beta$, $\delta$ and $r$, and $(\diamondsuit)$, $(\heartsuit)$, $(\spadesuit)$  with constants $\gamma$, $\nu$, $r$ and $\delta$.
\end{Definition}
In other words, if $H \subset \bar{F}$ contains $f^*$, then on an event with probability at least $1-\delta/2$, the triplet $(H,X,Y)$ satisfies $(\clubsuit)$, $(\diamondsuit)$, $(\heartsuit)$ and $(\spadesuit)$ with the same constants $\alpha$, $\beta$, $\gamma$, $\nu$, $\delta$ and $r$. To simplify notation we avoid repeating this long list of constants in what follows, and just use the term ``suitable" instead. Having said that, the list of constants should not be intimidating. As we show in what follows, in the context we focus on here, i.e., when $F \subset M B(L_p)$ and $\|Y\|_{L_p} \leq M$, the constants $\alpha$, $\beta$, $\gamma$ and $\nu$ turn out to be just absolute constants, and the two that really matter are the accuracy parameter $\eps=r^2$ and the confidence parameter $\delta$.

\vskip0.3cm

With those definitions in place, let us select the tuning parameters $\theta_2$, $\theta_3$ and $\theta_4$ accordingly, by stating how they depend on $\alpha$, $\beta$, $\gamma$ and $\nu$. One may select
$$
\theta_2= \frac{\beta^2}{\alpha^2} + \gamma; \ \ \ \theta_3 = \frac{2\nu}{\alpha^2}; \ \ \ \theta_4 =\beta,
$$
and the following fact was established in \cite{MenACM2}:
\begin{Theorem} \label{thm:prop-to-error}
There exists an absolute constant $c$ for which the following holds. Let $(F,X,Y)$ be a suitable triplet. Set
\begin{equation} \label{eq:in-thm-prop-to-error}
\rho=2\nu\left(1+\frac{\beta^2}{\alpha^2}\right) \ \ {\rm and} \ \ \bar{r}=\sqrt{2}\left(\gamma+\frac{\beta^2}{\alpha^2}\right)^{1/2} r.
\end{equation}
If $\rho \leq 1/18$ then the procedure $\hat{f}$ satisfies that with probability at least $1-\delta$,
$$
\E {\cal L}_{\hat{f}} \leq c\bar{r}^2.
$$
\end{Theorem}
Invoking Theorem \ref{thm:prop-to-error}, Theorem \ref{thm:main} can be established once one shows that if $F \subset MB(L_p)$ and $\|Y\|_{L_p} \leq M$, then the triplet $(F,X,Y)$ is suitable for $r=\sqrt{\eps}$ and $\delta$; $\alpha$, $\beta$, $\gamma$ and $\nu$ are absolute constants; and the resulting $\rho$ from \eqref{eq:in-thm-prop-to-error} which satisfies that $\rho \leq 1/18$. Also, one has to specify the final tuning parameter $\theta_1$ which will also be an absolute constant.

Let us now explore the types of estimates one should establish when verifying $(\clubsuit)$ and $(\diamondsuit)$. As we explain in what follows, $(\heartsuit)$ and $(\spadesuit)$ are based on a different argument (see Section \ref{sec:proof-on-main-end}).

\section{Towards $(\clubsuit)$ and $(\diamondsuit)$: Three features of a random variable}
Recall that the properties $(\clubsuit)$ and $(\diamondsuit)$ are of a similar flavour: ``guesses" of the $L_2$ norm of differences of class members that are valid uniformly. Thus, as a first step, let us study simpler versions of the two properties: consider a single random variable $W$. To establish $(\clubsuit)$ one has to be able to ``guess" $\|W\|_{L_2}$ in an isomorphic way from a sample $W_1,...W_N$, and for $(\diamondsuit)$ one has to obtain an almost isometric lower bound on $\|W\|_{L_2}^2$ using a majority vote of the empirical means $m^{-1}\sum_{i \in I_j} W_i^2$. Of course, one has to keep in mind even when considering a single random variable that $(\clubsuit)$ and $(\diamondsuit)$ have to be established uniformly in a class.

\vskip0.3cm

The basic phenomenon behind $(\clubsuit)$ and $(\diamondsuit)$ is that a sample can only ``see" events that are nontrivial. For example, if $W$ is a random variable and
$$
Pr \left( \alpha \|W\|_{L_2} \leq |W| \leq \beta \|W\|_{L_2}\right) \geq \delta,
$$
then with very high probability---at least $1-2\exp(-c\delta N)$, more than $\delta N/2$ of the sample points $W_1,...,W_N$ satisfy that
$$
|W_i| \in \left[\alpha \|W\|_{L_2}, \beta  \|W\|_{L_2} \right].
$$

One can easily ensure that $Pr( |W| \leq \beta \|W\|_{L_2})$ is large by invoking Chebyshev's inequality and taking $\beta$ sufficiently large. However, the lower estimate depends on $W$ satisfying a small-ball condition: not too much ``weight" is assigned close to $0$ (see Definition \ref{def:standard-assumption}). Since the small-ball condition is an intrinsic property of the random variable, one has no control on the values of the small-ball constants $\kappa$ and $\delta$. At times, that lack of flexibility is restrictive, as one would like $\delta$ to be close to $1$ with $\kappa$ sufficiently far from $0$.

One example of a random variable that has such a good small-ball behaviour is a centred gaussian random variable $g$: it satisfies that for any $\eta>0$,
$$
Pr\left (|g| \leq c\eta \|g\|_{L_2}\right) \leq \eta
$$
where $c$ is a suitable absolute constant.

As it happens, the proof of $(\clubsuit)$ calls for the creation of a gaussian-like behaviour (at least for every $\eta \geq \eta_0$ for a constant $\eta_0$ that can be made as small as one wishes). The obvious solution, using  that $\frac{1}{\sqrt{m}} \sum_{i=1}^m W_i$ is ``close" to a gaussian (a Berry-Esseen type argument), is only effective when $W$ satisfies some norm equivalence. Unfortunately, in the context of learning in $L_p$ no useful norm equivalence exists, and example \ref{eq:no-SB-L-infty} shows how bad the situation can be in such a case.

\vskip0.3cm

At the same time, the proof of $(\diamondsuit)$ requires one to show that, with high probability, $m^{-1}\sum_{i=1}^m W_i^2$ is an almost isometric lower bound on $\|W\|_{L_2}$. And, since there is a need for a uniform estimate in a class, it turns out that one has to ask for a little more: that the lower bound is stable under perturbations:

\begin{Definition} \label{def:stable-lower} \cite{MenStudia}
A random variable $W$ satisfies a stable lower bound with parameters $(\nu,\ell,k)$ for a sample size $m$ if the following holds. Let $W_1,...,W_m$ be independent copies of $W$. Then with probability at least $1-2\exp(-k)$, for any $J
\subset \{1,...,m\}$, $|J| \leq \ell$ one has that
$$
\frac{1}{m}\sum_{i \in J^c} W_i^2 \geq (1-\nu)\|W\|_{L_2}^2.
$$
\end{Definition}

The goal of this section is to show that both properties---a gaussian-like small-ball behaviour and a stable lower bound, can be established using an integrability condition: recall that the random variable $W$ satisfies a $(\Gamma,\xi)$ integrability condition if
$$
\E W^2 \IND_{\{|W| \geq \Gamma \|W\|_{L_2}\}} \leq \xi \E W^2.
$$

\begin{Theorem} \label{thm:integrability-implies-propoerties}
There exist absolute constants $c_0, c_1, c_2$ and $c_3$ for which the following holds. Let $W \in L_2$ that satisfies a $(\Gamma,\xi)$ integrability estimate for some $0<\xi \leq 1/100$.
\begin{description}
\item{$(1)$} For every integer $m$, $W$ satisfies a stable lower bound with parameters $(3\xi,\ell,\kappa)$ for
    $$
    \ell=c_0 m \frac{\xi }{\Gamma^2}, \ \ {\rm and} \ \ k = c_1 m \frac{\xi^2}{\Gamma^2}.
    $$

\item{$(2)$} If $0<\eta_0<1$ and $m \eta^2_0 \geq c_2\max\{1,\Gamma^2\}$, then for $\eta \geq \eta_0$,
$$
Pr \left( \left|\frac{1}{\sqrt{m}} \sum_{i=1}^m \eps_i W_i \right| \leq c_3\eta \|W\|_{L_2} \right) \leq \eta,
$$
where $(\eps_i)_{i=1}^m$ are independent, symmetric, $\{-1,1\}$-valued random variables that are independent of $W_1,...,W_m$.
\end{description}

\end{Theorem}

Theorem \ref{thm:integrability-implies-propoerties} is useful in the context of learning in $L_p$ because of the following simple observation:
\begin{Lemma} \label{lemma:L-p-integrability}
Let $p>2$. Every $W \in L_p$ satisfies a $(\Gamma,\xi)$ integrability condition for
$$
\Gamma = \left(\frac{\|W\|_{L_p}}{\|W\|_{L_2}}\right)^{\frac{p}{p-2}} \cdot \left(\frac{1}{\xi}\right)^{\frac{1}{p-2}}.
$$
\end{Lemma}

\proof Let $2r=p$ and set $r^\prime$ be the conjugate index of $r$. Then,
$$
\E W^2 \IND_{\{|W| \geq \Gamma \|W\|_{L_2}\}} \leq \left(\E W^{2r}\right)^{1/r} Pr^{\frac{1}{r^\prime}} \left(|W| \geq \Gamma \|W\|_{L_2}\right),
$$
and
$$
Pr^{\frac{1}{r^\prime}} \left(|W| \geq \Gamma \|W\|_{L_2}\right) \leq \left(\frac{\|W\|_{L_p}}{\Gamma \|W\|_{L_2}}\right)^{p-2}.
$$
\endproof

The proof of the first part of Theorem \ref{thm:integrability-implies-propoerties} can be found in \cite{MenStudia}. For the sake of completeness, let us sketch it.

\vskip0.3cm

\noindent{\bf Proof of Theorem \ref{thm:integrability-implies-propoerties}, part $(1)$---sketch.} Consider the `cutoff' level $M=\Gamma \|W\|_{L_2}$ and set $U=W\IND_{\{|W| \leq M\}}$.  The integrability condition implies that $\E U^2 \geq (1-\xi)\E W^2$. Therefore, it suffices to show that with high probability,
\begin{equation} \label{eq:U-cond-1}
\frac{1}{m} \sum_{i=1}^m U_i^2 \geq (1-\xi) \E U^2,
\end{equation}
and that for every $J \subset \{1,...,m\}$, $|J| \leq \ell$,
\begin{equation} \label{eq:U-cond-2}
\frac{1}{m} \sum_{j \in J} U_j^2 \leq \xi \E U^2;
\end{equation}
indeed, the combination of the two implies that
$$
\frac{1}{m} \sum_{j \in J^c} W_i^2 \geq \frac{1}{m} \sum_{j \in J^c} U_i^2 \geq (1-2\xi) \E U^2 \geq (1-3\xi)\E W^2.
$$
Equation \eqref{eq:U-cond-2} is straightforward with the choice of $\ell$, recalling that $\xi<1/100$ and that
$$
\|U\|_{L_\infty} \leq \Gamma \|W\|_{L_2} \leq \frac{\Gamma}{1-\xi} \|U\|_{L_2}.
$$
As for \eqref{eq:U-cond-1}, since $\|U\|_{L_\infty} \leq M$ it follows from Bernstein's inequality that
$$
Pr \left( \left|\frac{1}{m} \sum_{i=1}^m U_i - \E U^2\right| \geq t \right) \leq 2\exp\left(-cm \min\left\{\frac{t^2}{M^2 \E U^2}, \frac{t}{M^2}\right\} \right),
$$
and the wanted estimate on $k$ is evident by setting $t=\xi \E U^2$.
\endproof

The argument necessary for establishing the second part of Theorem \ref{thm:integrability-implies-propoerties} is based on Esseen's inequality.

\begin{Theorem} \label{thm:esseen}
There exists an absolute constant $c$ for which the following holds. Let $X$ be a random variable and set $F_X$ to be its characteristic function. Then for every $r>0$,
$$
\sup_{x \in \R} Pr \left( \left|X-x_0\right| \leq r \right) \leq c r \int_{-1/r}^{1/r} |F_X(t)|dt.
$$
\end{Theorem}

\vskip0.3cm

\noindent{\bf Proof of Theorem \ref{thm:integrability-implies-propoerties}---part $2$.} Let $X=\frac{1}{\sqrt{m}}\sum_{i=1}^m \eps_i W_i$. Since $\eps W$ is symmetric and $\eps$ is independent of $W$,
$$
F_X(t) = \left(\E \exp\left(it\left(\frac{\eps W}{\sqrt{m}}\right)\right)\right)^m = \left(\E \left(\cos\left(\frac{tW}{\sqrt{m}}\right)\right) \right)^m.
$$

Fix $t>0$ in a range that is specified in what follows. Set $z=1/2t$ and consider
$$
\cos\left(\frac{tW}{\sqrt{m}}\right) = \cos\left(\frac{W}{2z\sqrt{m}}\right) \IND_{\{|W| \leq \sqrt{m} z \}} + \cos\left(\frac{W}{2z\sqrt{m}}\right) \IND_{\{|W| > \sqrt{m} z \}} \equiv (1)+(2).
$$
To estimate $\E(1)$, observe that by the Taylor expansion of $\cos(x)$,
$$
\cos\left(\frac{W}{2z\sqrt{m}}\right) \IND_{\{|W| \leq \sqrt{m} z \}} = \left(1 - \frac{W^2}{8z^2m} + \sum_{j=2}^\infty (-1)^{j} \left(\frac{W}{2z\sqrt{m}}\right)^{2j}\frac{1}{(2j)!} \right) \IND_{\{|W| \leq \sqrt{m} z \}}.
$$
If $|W/\sqrt{m}|>z$ then $(1)=0$; otherwise $|W/(2z \sqrt{m})|\leq 1/2$ and let $a_j=(W/2z\sqrt{m})^{2j} \cdot (1/(2j)!)$. Clearly, $(a_j)_{j \geq 2}$ is nonnegative and decreasing, and in particular

$$
\left| \sum_{j=2}^\infty (-1)^{j} \left(\frac{W}{2z\sqrt{m}}\right)^{2j}\frac{1}{(2j)!} \right| \leq  a_2 =  \frac{W^4}{96z^4 m^2}  \leq \frac{W^2}{48z^2m},
$$
using once again that $|W/(2z\sqrt{m})| \leq 1/2$. Therefore, pointwise,
$$
(1) = \left(1 - \frac{W^2}{8z^2m} \left(1 + \bigtriangleup\right) \right) \IND_{\{|W| \leq \sqrt{m} z \}},
$$
with $|\bigtriangleup| \leq 1/12$ almost surely, implying that $(1)$ is nonnegative. Thus,
$$
0 \leq (1) \leq 1 - \frac{W^2}{6 z^2m} \left(1 - \IND_{\{|W| > \sqrt{m} z \}}\right)
$$
and
$$
\E (1) \leq 1 - \frac{1}{6 z^2 m}\left(\E W^2 - \E W^2 \IND_{\{|W|> z\sqrt{m}\}}\right) \leq 1 - \frac{1}{6} \frac{\E W^2}{z^2 m}\cdot  (1-\xi),
$$
where the last inequality follows from the integrability condition as long as
\begin{equation} \label{eq:condition-on-z-1}
\frac{z\sqrt{m}}{\|W\|_{L_2}} \geq \Gamma.
\end{equation}

At the same time, again if \eqref{eq:condition-on-z-1} holds, one has that
\begin{align*}
& \E \left|\cos\left(\frac{W}{2z\sqrt{m}}\right) \IND_{\{|W| > \sqrt{m} z \}}\right| \leq Pr \left(|W| > \sqrt{m} z\right)
\\
= & Pr \left(|W|\IND_{\{|W|>\sqrt{m}z\}} > \sqrt{m} z\right) \leq \frac{\E W^2 \IND_{\{|W|>\sqrt{m}z\}}}{z^2m} \leq \xi \frac{ \E W^2}{z^2 m}.
\end{align*}
Therefore, recalling that $\xi \leq 1/100$ and that $t=1/2z$,
$$
\left|\E \cos\left(\frac{W}{2z^2\sqrt{m}}\right) \right| \leq 1 - \frac{\E W^2}{z^2m} \left(\frac{1}{6} (1-\xi)-\xi\right) \leq 1 - \frac{\E W^2}{8z^2m}=1-\frac{t^2 \E W^2}{2m}
$$
provided that
\begin{equation} \label{eq:condition-on-m-1}
m \geq c_0 t^2 \E W^2
\end{equation}
and in which case,
$$
|F_X(t)| \leq \left(1-\frac{t^2\E W^2}{2m}\right)^m \leq \exp(-c_1 t^2 \E W^2),
$$
for suitable absolute constants $c_0$ and $c_1$.

Now let us consider $r = \eta \|W\|_{L_2}$, implying that integration in Esseen's inequality (Theorem \ref{thm:esseen}) takes place in $\left[-\frac{1}{\eta \|W\|_{L_2}}, \frac{1}{\eta \|W\|_{L_2}} \right]$. Therefore, $|t| \leq 1/\eta \|W\|_{L_2}$,  \eqref{eq:condition-on-z-1} holds if $\eta^2 m \gtrsim \Gamma^2$ and \eqref{eq:condition-on-m-1} is verified if $\eta^2 m \gtrsim 1$. By the lower bound on $m$ in the assumption, the estimate on $|F_X(t)|$ holds for every $t$ in the interval $[-1/r,1/r]$, and by Esseen's Theorem and a change of variables,
$$
Pr \left( |X| \leq \eta \|W\|_{L_2} \right) \leq cr \int_{[-1/r,1/r]} \exp(-c_0 t^2 \E W^2) dt \leq c_2 \eta,
$$
as claimed.
\endproof

\section{Towards $(\clubsuit)$ and $(\diamondsuit)$ --- Uniform estimates}
For a function $h \in L_2(\mu)$ denote by $\Gamma(h,\xi)$ the integrability constant
$$
\Gamma(h,\xi) = \inf \left\{ \Gamma : \E h^2 \IND_{\{|h| \geq \Gamma \|h\|_{L_2}\}} \leq \xi \E h^2 \right\},
$$
and observe that for every $h$, $\Gamma(h,\xi) \geq 1-\xi$; in particular, $\Gamma(h,1/100) \geq 1/2$.

Let $H \subset L_2(\mu)$ be a class of functions that is star-shaped around $0$ and set $\Gamma(r)$ that satisfies $$
\sup_{u \in (H-H) \cap rS(L_2)} \Gamma(u,1/100) \leq \Gamma(r),
$$
where throughout this note, $r S(L_2)=\{f : \|f\|_{L_2} =r\}$.

\begin{Remark}
The star-shape property of $H$ implies that the sets $\{h/\|h\|_{L_2} : \|h\|_{L_2} = r\}$ become larger the smaller $r$ is. Thus, one may assume that $\Gamma(r)$ is decreasing. Moreover, let us stress that there need not be a uniform estimate on $\Gamma(r)$; it may deteriorate as $r$ goes to $0$.
\end{Remark}

Before formulating the wanted uniform estimate, one requires an additional preliminary result: a Sudakov type bound. To that end, let ${\cal M}(H,\eps D)$ be the maximal cardinality of a subset of $H$ that is $\eps$-separated with respect to the $L_2(\mu)$ norm.

\begin{Theorem} \label{thm:Sudakov-implies-covering} \cite{MenStudia}
For any $0<\theta \leq 1/2$ and $\eta >0$ there is a constant $c(\theta,\eta)$ for which the following holds: let $r>0$ and set
$$
\sup_{u \in (H-H) \cap \theta r S(L_2)} \Gamma(u,1/100) \leq \Gamma.
$$
If
$$
\log {\cal M}(H \cap r S(L_2), \theta r D) \geq \eta \frac{N}{\Gamma^2}
$$
then
$$
\E \sup_{h \in H \cap r S(L_2)} \left|\frac{1}{N} \sum_{i=1}^N \eps_i h(X_i) \right| \geq c(\theta,\eta) \frac{r}{\Gamma}.
$$
\end{Theorem}

In the case that interests us, both $\theta$ and $\eta$ turn out to be just absolute constants, and therefore so is $c(\theta,\eta)$. As a result, for any given values of $\theta$ and $\eta$, the fact that
\begin{equation} \label{eq:fixed-point-in-iso-1}
\E \sup_{h \in H \cap r D} \left|\frac{1}{N} \sum_{i=1}^N \eps_i h(X_i) \right| \leq \frac{c(\theta,\eta)}{2} \frac{r}{\Gamma},
\end{equation}
implies that
$$
\log {\cal M}(H \cap r S(L_2), \theta r D) \leq \eta \frac{N}{\Gamma^2}.
$$

\begin{Example} \label{ex:L-p-integrability}
Let $H \subset M B(L_p(\mu))$. Then $H-H \subset 2M B(L_p(\mu))$, and any $u \in (H-H) \cap \theta r S(L_2)$ satisfies a $(\Gamma,1/100)$ integrability condition for $\Gamma=c(M/\theta r)^{p/(p-2)}$. In particular, given $0<\theta<1/2$ and $\eta>0$, if
$$
\E \sup_{h \in H \cap r S(L_2)} \left|\frac{1}{N}\sum_{i=1}^N \eps_i h(X_i) \right| \leq \frac{c(\theta,\eta)}{2} \cdot r \left(\frac{\theta r}{M}\right)^{\frac{p}{p-2}},
$$
one has that
$$
\log {\cal M} \left( H \cap r S(L_2), \theta r D\right) \leq N \left(\frac{\theta r}{M}\right)^{\frac{2p}{p-2}}.
$$
\end{Example}

\subsection{ $(\clubsuit)$---an $L_2$ norm oracle}
As is the case throughout this section, let $H \subset L_2(\mu)$ be star-shaped around $0$. For $r>0$, let $\Gamma(r)$ be a decreasing function which satisfies that
$$
\sup_{u \in (H-H) \cap r S(L_2)} \Gamma(u,1/100) \leq \Gamma(r).
$$
Consider absolute constants $c_0$ and $c_1<1$ and fix $r>0$ such that
\begin{equation} \label{eq:distance-oracle-fixed-point-1}
\E \sup_{h \in H \cap rD} \left|\frac{1}{N}\sum_{i=1}^N \eps_i h(X_i) \right| \leq c_0 \frac{r}{\Gamma^2(c_1r)}.
\end{equation}
Set
$$
\Gamma(r) \lesssim m \lesssim \Gamma(c_1 r) \ \ {\rm and} \ \ n = \frac{N}{m},
$$
let $\sigma=(X_i)_{i=1}^N$ and set $I_1,...,I_n$ to be a partition of $\{1,...,N\}$ to blocks of cardinality $m$. Put
$$
\mu_j(h) =  \left|\frac{1}{\sqrt{m}} \sum_{i \in I_j} \eps_i h(X_i)\right|, \ \ j=1,...,n,
$$
and define
$$
\Psi(h,\sigma)= {\rm Med} \left\{\mu_1(h),...,\mu_n(h)\right\}
$$
to be a median of $(\mu_j(h))_{i=1}^n$.

\begin{Theorem} \label{thm:iso-distance}
There are absolute constants $c$, $c^\prime$, $c^{\prime \prime}$, $\alpha$ and $\beta$ such that the following holds. Let $H$ be star-shaped around $0$ and let $r$ satisfy \eqref{eq:distance-oracle-fixed-point-1}. Then with probability at least $1-2\exp(-cn) \geq 1-2\exp(-c^\prime N/\Gamma^2(c^{\prime \prime} r))$, for every $h \in H$ one has that
\begin{description}
\item{$\bullet$} If $\Psi(h,\sigma) > \beta r$ then $\beta^{-1} \Psi(h,\sigma) \leq \|h\|_{L_2} \leq \alpha^{-1} \Psi(h,\sigma)$.
\item{$\bullet$} If $\Psi(h,\sigma) \leq \beta r$ then  $\|h\|_{L_2} \leq (\beta/\alpha) r$.
\end{description}
\end{Theorem}

\begin{Remark}
In the context of a distance oracle, given a triplet $(U,X,Y)$, set $H={\rm star}(U-u^*,0)$. Thus, a norm oracle for $H$ can be used to estimate the $L_2(\mu)$ distances between functions in $U$ and $u^*$, as the learning procedure requires.
\end{Remark}

The first step in the proof of Theorem \ref{thm:iso-distance} is a simple observation:
\begin{Lemma} \label{lemma:ok-for-distantce-oracle}
Assume that there are constants $c$ and $C$, $r>0$ and an event ${\cal A}$ on which one has that
\begin{description}
\item{$(1)$} If $\|h\|_{L_2} > r$ then $c \|h\|_{L_2} \leq \Psi(h,\sigma) \leq C\|h\|_{L_2}$.
\item{$(2)$} If $\|h\|_{L_2} \leq r$ then $\Psi(h,\sigma) \leq Cr$.
\end{description}
Then on the event ${\cal A}$, $\Psi$ satisfies the assertion of Theorem \ref{thm:iso-distance} with constants $\alpha=c$ and $\beta=C$.
\end{Lemma}

\proof
Let $\sigma \in {\cal A}$. If $\Psi(h,\sigma) > Cr$ then by $(2)$, $\|h\|_{L_2} > r$; hence, by $(1)$,
\begin{equation} \label{eq:two-sided-psi}
C^{-1} \Psi(h,\sigma) \leq \|h\|_{L_2} \leq c^{-1} \Psi(h,\sigma).
\end{equation}
Otherwise, $\Psi(h,\sigma) \leq Cr$. If, in addition, $\|h\|_{L_2} > r$, then, again, it follows from \eqref{eq:two-sided-psi} that  $\|h\|_{L_2} \leq c^{-1} \Psi(h,\sigma) \leq (C/c) r$. And if $\|h\|_{L_2} \leq r$ then in particular
$\|h\|_{L_2} \leq (C/c) r$, as claimed.
\endproof

\vskip0.3cm

\noindent{\bf Proof of Theorem \ref{thm:iso-distance}.} The proof follows by showing that with probability at least $1-2\exp(-c_0 n)$, for every $h \in H$, $\Psi(h,\sigma)$ satisfies properties $(1)$ and $(2)$ of Lemma \ref{lemma:ok-for-distantce-oracle}, and for the stated value of $r$. The probability estimate is evident because $n =N/m$ and using the upper bound on the choice of $m$.

Fix $h \in H \cap rS(L_2)$ and recall that
$$
\mu_j(h) = \left|\frac{1}{\sqrt{m}} \sum_{i \in I_j} \eps_i h(X_i) \right|, \ \ j=1,...,n.
$$
Thus, $(\mu_j(h))_{j=1}^n$ are $n$ independent copies of $Y=\left|m^{-1/2}\sum_{i=1}^m \eps_i h(X_i)\right|$. Clearly, $\E Y^2 = \E h^2=r^2$. Moreover, by the second part of Theorem \ref{thm:integrability-implies-propoerties} for $\eta_0=0.01$, it follows that if $m \geq c_1 \max\{1,\Gamma^2(r)\} \geq c_1^\prime \Gamma^2(r)$, then
$$
Pr \left( |Y| \geq c_2\|h\|_{L_2} \right) \geq 0.99,
$$
for an absolute constant $c_2$. The upper estimate on $|Y|$ is an immediate outcome of Chebyshev's inequality, implying that there is an absolute constant $c_3$ such that
$$
Pr \left( |Y| \in \left[c_2\|h\|_{L_2},c_3\|h\|_{L_2}\right] \right) \geq 0.98.
$$
Hence, by a binomial estimate, with probability at least $1-2\exp(-c_4n)$
$$
\left|\left\{j : \mu_j(h) \in \left[c_2\|h\|_{L_2},c_3\|h\|_{L_2}\right] \right\} \right| \geq 0.97 n.
$$

Now let $V \subset H \cap r S(L_2)$ be a maximal separated set of cardinality  $\exp(c_4 n/2)$ and denote its mesh width by $\rho$. 
For every $h \in H \cap r S(L_2)$ let $\pi h$ be the nearest point to $h$ in $V$ with respect to the $L_2(\mu)$ norm. Let us show that with high probability,
\begin{equation} \label{eq:osc-distance-proof}
(*)=\sup_{h \in H \cap r S(L_2)} \left| \left\{j : |\mu_j(h)-\mu_j(\pi h)| \leq (c_2/2)r \right\} \right| \leq 0.03n.
\end{equation}
In that case, for any $h \in H \cap rS(L_2)$, for at least $0.94n$ indices $j$,
$$
\frac{c_2}{2} \|h\|_{L_2} \leq \mu_j(h) \leq \left(c_3+\frac{c_2}{2}\right) \|h\|_{L_2}.
$$

To establish \eqref{eq:osc-distance-proof}, note that by linearity
$$
\left|\mu_j(h) - \mu_j(\pi h)\right| = \left| \frac{1}{\sqrt{m}} \sum_{i \in I_j} \eps_i (h-\pi h)(X_i) \right|=\left|\mu_j(h-\pi h)\right|,
$$
and the random variable
$$
Z = \sup_{h \in H \cap r S(L_2)} \sum_{j=1}^n \IND_{\{|\mu_j(h-\pi h)| \geq c_2r/2\}}
$$
concentrates well around its mean. Indeed, by the bounded differences inequality (see, e.g. \cite{BoLuMa13}), with probability at least $1-2\exp(-c_5 n)$,
$$
Z \leq \E Z + 0.01n.
$$
To control $\E Z$, note that
\begin{align*}
\E Z \leq & \frac{2}{c_2 r} \E \sup_{h \in H \cap r S(L_2)} \sum_{j=1}^n  \left(\left| \frac{1}{\sqrt{m}} \sum_{i \in I_j} \eps_i (h-\pi h)(X_i)\right| - \E \left| \frac{1}{\sqrt{m}} \sum_{i \in I_j} \eps_i (h-\pi h)(X_i)\right| \right)
\\
+ & \frac{2n}{c_2r} \E  \left|\frac{1}{\sqrt{m}} \sum_{i=1}^m \eps_i (h -\pi h)(X_i) \right| = (1)+(2).
\end{align*}
Clearly,
$$
(2) \leq \frac{2n}{c_2r} \|h-\pi h\|_{L_2} \leq \frac{2n \rho}{c_2 r} \leq 0.01n
$$
provided that $\rho \leq c_6r$ for a suitable absolute constant $c_6$. Thus, to ensure that the constraint on the cardinality of $V$ is satisfied, one has to verify that
\begin{equation} \label{eq:BE-entropy-1}
\log {\cal M}\left(H \cap rS(L_2), c_6r\right) \leq \frac{c_4}{2} n = \frac{c_4}{2} \frac{N}{m}.
\end{equation}

Moreover, by a standard symmetrization and contraction argument, with $(\eps_j^\prime)$ that are independent of $(\eps_i)_{i=1}^N$,
\begin{align*}
& \E \sup_{h \in H \cap r S(L_2)} \left| \sum_{j=1}^n \left( \left| \frac{1}{\sqrt{m}} \sum_{i \in I_j} \eps_i (h-\pi h)(X_i) \right| - \E \left| \frac{1}{\sqrt{m}} \sum_{i \in I_j} \eps_i (h-\pi h)(X_i) \right| \right) \right|
\\
\leq & 2 \E \sup_{h \in H \cap r S(L_2)} \left| \sum_{j=1}^n  \eps_j^\prime \cdot \left| \frac{1}{\sqrt{m}} \sum_{i \in I_j} \eps_i (h-\pi h)(X_i) \right| \right|
\\
\leq & 2 \E \sup_{h \in H \cap r S(L_2)} \left| \sum_{j=1}^n  \eps_j \cdot \left( \frac{1}{\sqrt{m}} \sum_{i \in I_j} \eps_i (h-\pi h)(X_i) \right) \right|
\\
\leq & \frac{4}{\sqrt{m}} \E \sup_{h \in H \cap r S(L_2)} \left|\sum_{i=1}^N \eps_i (h-\pi h)(X_i) \right|  =(3),
\end{align*}
and one has to ensure that $(3) \leq c_8rn$. Therefore, by the triangle inequality it suffices that
\begin{equation} \label{eq:BE-Rad-1}
\E \sup_{h \in H \cap r D} \left|\frac{1}{N} \sum_{i=1}^N \eps_i h(X_i) \right| \leq c_9 \frac{r}{\sqrt{m}}.
\end{equation}

Recall Theorem \ref{thm:Sudakov-implies-covering} and its notation. Set $\theta=c_6$, and thus, to use Theorem \ref{thm:Sudakov-implies-covering}, one requires that
$$
\frac{c_4}{2} \cdot \frac{N}{m} \geq \frac{\eta N}{\Gamma^2(\theta r)},
$$
i.e., that $m \lesssim \eta^{-1} \Gamma^2(\theta r)$ and that
$$
\E \sup_{h \in H \cap r D} \left|\frac{1}{N} \sum_{i=1}^N \eps_i h(X_i) \right| \leq c(\theta,\eta) \frac{r}{\Gamma^2(\theta r)}.
$$
Therefore, fix $\eta$ to be an absolute constant, set $c_{10}=c(\theta,\eta)$ and $c_{11}=\frac{1}{2}\min\{c_9 \sqrt{\eta}, c_{10}\}$. Recall that $\theta=c_6$. It follows from  Theorem \ref{thm:Sudakov-implies-covering} and the fact that $H$ is star-shaped around $0$ that if
$r$ satisfies that
$$
\E \sup_{h \in H \cap r D} \left|\frac{1}{N} \sum_{i=1}^N \eps_i h(X_i) \right| \leq c_{11} \frac{r}{\Gamma^2(c_6 r)},
$$
then both \eqref{eq:BE-entropy-1} and \eqref{eq:BE-Rad-1} hold for that value or $r$.

Hence, with probability at least
$$
1-2\exp(-c_5n) \geq 1-2\exp(-c^\prime N/\Gamma^2(c^{\prime \prime} r)),
$$
one has that
 $\Psi(h,\sigma) \sim \|h\|_{L_2}$ for any $h \in H \cap rS(L_2)$. By homogeneity and the fact that $H$ is star-shaped around $0$, the same holds for any $h \in H$ which satisfies that $\|h\|_{L_2} \geq r$, as required.

Finally,  the case $\|h\|_{L_2} \leq r$, in which only an upper estimate is required, can be verified using a similar argument to the one used in $H \cap r S(L_2)$---namely, by showing that with high probability,
$$
\sup_{h \in H \cap r D} \sum_{j=1}^n \IND_{\{|\mu_j(h)| \geq c r \}} \leq 0.94n.
$$
We omit the details of the proof.
\endproof

Let us formulate the estimate for $(\clubsuit)$ in the learning procedure, which is an immediate outcome of Theorem \ref{thm:iso-distance}, by setting $H={\rm star}(U-u^*,0)$ and recalling that when $U \subset MB(L_p(\mu))$ one may take $\Gamma(r)=c(M/r)^{p/(p-2)}$ for a suitable absolute constant $c$.
\begin{framed}
\begin{Corollary}  \label{cor:distance-oracle-L-p}
There are absolute constants $c_0$, $c_1$, $\alpha$ and $\beta$ for which the following holds.

Let $(U,X,Y)$ be a triplet where $U \subset M B(L_p(\mu))$. Set $r>0$ such that
\begin{equation} \label{eq:in-cor-distance-oracle-L-p}
\E \sup_{u \in U_{f^*,r}} \left|\frac{1}{N}\sum_{i=1}^N \eps_i u(X_i) \right| \leq c_0 r \left(\frac{r}{M}\right)^{\frac{p}{p-2}},
\end{equation}
where, as always, $U_{u^*,r}={\rm star}(U-u^*,0) \cap r D$. Consider $\Psi$ as in Theorem \ref{thm:iso-distance} for $m \sim (M^2/r^2)^{p/(p-2)}$. Then with probability at least $1-2\exp(-c_1 N (r^2/M^2)^{p/(p-2)})$, for every $u \in U$ one has
\begin{description}
\item{$\bullet$} If $\Psi(u-u^*,\sigma) > \beta r$ then $\beta^{-1} \Psi(u-u^*,\sigma) \leq \|u-u^*\|_{L_2} \leq \alpha^{-1} \Psi(u-u^*,\sigma)$;
\item{$\bullet$} If $\Psi(u-u^*,\sigma) \leq \beta r$ then  $\|u-u^*\|_{L_2} \leq (\beta/\alpha) r$.
\end{description}
\end{Corollary}
In particular, if $N_0$ is as in Theorem \ref{thm:main} and $N \geq N_0$, then $r=\sqrt{\eps}$ satisfies \eqref{eq:in-cor-distance-oracle-L-p} and
$1-2\exp(-c_1 N (r^2/M^2)^{p/(p-2)}) \geq 1-\delta/4$. Thus, $(\clubsuit)$ holds for any such triplet for $\alpha$ and $\beta$ that are absolute constants.
\end{framed}

\begin{Remark} \label{rem:value-of-nu}
Thanks to Theorem \ref{thm:iso-distance}, the constants $\alpha$ and $\beta$ used in the learning procedure are just well-chosen absolute constants. Using the notation of Theorem \ref{thm:prop-to-error}, all one has to ensure is that $\nu$ is sufficiently small to imply that
$$
\rho=2 \nu \left(1+\frac{\beta^2}{\alpha^2}\right) \leq \frac{1}{18}.
$$
Thus, $\nu$ is simply an absolute constant.
\end{Remark}

\subsection{$(\diamondsuit)$ --- An almost isometric lower bound}
Recall that $H$ is star-shaped around $0$.  Consider a function $\Gamma(r,\xi)$ which satisfies that
$$
\sup_{u \in (H-H) \cap rS(L_2)} \Gamma(u,\xi) \leq \Gamma(r,\xi)
$$
and is decreasing as a function of $r$ for any fixed $\xi$ and decreasing in $\xi$ for any fixed $r$.

Property $(\diamondsuit)$ calls for a uniform, almost isometric lower bound on
$$
\frac{1}{m}\sum_{i \in I_j} h^2(X_i)
$$
that holds for most blocks $I_j$. The wanted estimate is based on the ``moreover" part of Theorem~3.1 from \cite{MenStudia}, with the choice of $\eta=0.01$ in its formulation there.

\begin{Theorem} \label{thm:main-lower}
There exist absolute constants $c_0,c_1,c_2$ and $c_3$ for which the following holds.  Let $H$ be star-shaped around $0$, fix $0<\xi<1$ and let $r>0$ such that
\begin{description}
\item{$(1)$} Every $h \in H \cap rS(L_2)$ satisfies a stable lower bound with parameters
    $(\xi/2,\ell,k)$, for $k \geq c_0$.
\item{$(2)$} $\E \sup_{u \in H \cap rD} \left|\frac{1}{N} \sum_{i=1}^N \eps_i u(X_i)
    \right| \leq c_1 \xi r
    \cdot\min\left\{\sqrt{\frac{\ell}{m}},\sqrt{\frac{k}{m}}\right\}$.
\item{(3)} If $h_1,h_2 \in H \cap rS$ and $\|h_1-h_2\|_{L_2} \geq c_2
    \xi r$ then $h_1-h_2$ satisfies a stable lower bound with parameters
    $(1/2,\ell,k)$.
\end{description}
Then with probability at least
$$
1-2\exp\left(-c_3 N \min\left\{\frac{\ell}{m},\frac{k}{m}\right\}\right)
$$
one has that
$$
\inf_{\{h \in H \ : \ \|h\|_{L_2} \geq r\}} \left| \left\{ j : \frac{1}{m}\sum_{i \in
I_j} h^2(X_i) \geq (1-\xi)\|h\|_{L_2}^2 \right\} \right| \geq 0.99n.
$$
\end{Theorem}

Since $0 \in H$ (the set $H$ is star-shaped around $0$) and $(H-H) \cap rS(L_2)$ satisfies the integrability condition, it is evident that the stable lower bound required for $(1)$ in Theorem \ref{thm:main-lower} holds for
$$
\ell = c m\frac{\xi}{\Gamma^2(r,\xi/6)} \ \ {\rm and} \ \ k = c^\prime m \frac{\xi^2}{\Gamma^2(r,\xi/6)}.
$$
At the same time, the stable lower bound required for $(3)$ holds with constants
$$
\ell = c \frac{m}{\Gamma^2(c^{\prime \prime} \xi r,1/12)} \ \ {\rm and} \ \ k = c^\prime \frac{m}{\Gamma^2(c^{\prime \prime} \xi r,1/12)},
$$
with $c$, $c^\prime$ and $c^{\prime \prime}$ are absolute constants.

Set $\xi=\min\{\nu,1/100\}$. By Remark \ref{rem:value-of-nu}, $\nu$ is just an absolute constant and therefore so is $\xi$. By the monotonicity properties of $\Gamma(r,\xi)$ one may set
$$
\ell = c_0 \frac{m}{\Gamma^2(c_1 r,c_2)} \ \ {\rm and} \ \ k = c_0^\prime \frac{m}{\Gamma^2(c_1^\prime r,c_2^\prime)}.
$$
for well-chosen absolute constants. Thus, one has the following:

\begin{Theorem} \label{thm:almost-isomtric}
There exist absolute constants $c_0,...,c_5$ for which the following holds. Let $r$ satisfy that
$$
\E \sup_{h \in H \cap rD} \left|\frac{1}{N} \sum_{i=1}^N \eps_i h(X_i).
    \right| \leq c_0 \frac{r}{\Gamma^2(c_1 r,c_2)},
$$
Then with probability at least $1-2\exp(-c_3 N/\Gamma^2(c_4 r,c_5))$, for every $h \in H$ such that $\|h\|_{L_2} \geq r$ one has
\begin{equation} \label{eq:in-thm-almost-iso-1}
\left| \left\{ j : \frac{1}{m}\sum_{i \in
I_j} h^2(X_i) \geq (1-\nu)\|h\|_{L_2}^2 \right\} \right| \geq 0.99n,
\end{equation}
where $\nu$ is the absolute constant from Remark \ref{rem:value-of-nu}.
\end{Theorem}

\vskip0.3cm

Let us formulate the outcome of Theorem \ref{thm:almost-isomtric} in the context of learning in $L_p$, setting, once again, $H={\rm star}(U-u^*,0)$ and recalling that since $U \subset MB(L_p(\mu))$ one may set $\Gamma(r,\xi)=c(M/r)^{p/(p-2)} \cdot (1/\xi)^{1/(p-2)}$.

\begin{framed}
\begin{Corollary} \label{cor:almost-iso-L-p}
There are absolute constants $c_0$ and $c_1$ for which the following holds. Let $(U,X,Y)$ be a triplet and assume that $U \subset M B(L_p(\mu))$ for some $p>4$. Set $r$ such that
\begin{equation} \label{eq:in-cor-almost-iso-L-p}
\E \sup_{u \in U_{f^*,r}} \left|\frac{1}{N} \sum_{i=1}^N \eps_i u(X_i) \right| \leq c_0 r \left(\frac{r}{M}\right)^{\frac{p}{p-2}}.
\end{equation}
Then with probability at least $1-2\exp(-c_1 N (r^2/M^2)^{p/(p-2)})$, if $u \in U$ and $\|u-u^*\|_{L_2} \geq r$ then
$$
\left| \left\{ j : \frac{1}{m}\sum_{i \in
I_j} (u-u^*)^2(X_i) \geq (1-\nu)\|u-u^*\|_{L_2}^2 \right\} \right| \geq 0.99n,
$$
where $\nu$ is the absolute constant from Remark \ref{rem:value-of-nu}.
\end{Corollary}
In particular, if $N_0$ is as in Theorem \ref{thm:main} and $N \geq N_0$ then $r=\sqrt{\eps}$ satisfies \eqref{eq:in-cor-almost-iso-L-p}, and the assertion of Corollary \ref{cor:almost-iso-L-p} holds with probability at least $1-\delta/4$. Thus, every such triplet satisfies $(\diamondsuit)$ as required.
\end{framed}

\section{$(\heartsuit)$, $(\spadesuit$)--- and the proof of Theorem \ref{thm:main}} \label{sec:proof-on-main-end}
The final component in the proof of Theorem \ref{thm:main} is showing that $(\heartsuit)$ and $(\spadesuit)$ hold, by studying the behaviour of the centred multiplier process. As it happens, the wanted estimate has been established in \cite{MenACM2}, and we sketch the argument for the sake of completeness, modifying it to fit the case when $U \subset M B(L_p(\mu))$ and $\|Y\|_{L_p} \leq M$.

Recall that $m$ is of the order of $(M^2/r^2)^{p/(p-2)}$. Given a triplet $(U,X,Y)$ let $u^* \in {\rm argmin}_{u \in U} (u(X)-Y)^2$, set $\xi=u^*(X)-Y$, and in particular $\|\xi\|_{L_p} \leq 2M$. Assume further that $r$ satisfies
$$
\E \sup_{u \in U_{u^*,r}} \left|\frac{1}{N}\sum_{i=1}^N \eps_i \xi_i u(X_i)\right| \leq c \left(\frac{r}{M}\right)^{p/(p-2)} r,
$$
for a well-chosen absolute constant $c$.

\vskip0.3cm
The first observation is straightforward:

\begin{Lemma} \label{lemma:L-p-multiplier}
Let $p > 4$ and assume that $\|\xi\|_{L_p}, \|h\|_{L_p} \leq M$. Then
$$
\|\xi h\|_{L_2} \leq c M^{\frac{p}{p-2}} \|h\|_{L_2}^{1-\frac{2}{p-2}}
$$
where $c$ is an absolute constant.
\end{Lemma}

\proof
Set $0<\alpha<1$ and observe that
$$
|\xi h|^2 \leq |h|^{2\alpha} (|h|^{2-2\alpha} |\xi|^2) \leq |h|^{2\alpha} (\max\{|h|, |\xi|\})^{2(2+\alpha)}.
$$
Therefore, it suffices to estimate $\E |h|^{2\alpha} |u|^{2(2+\alpha)}$, where $\|u\|_{L_p} \leq 2M$. Set $\alpha=(p-4)/(p-2)$, $q=(p-2)/2$, and put $q^\prime$ to be the conjugate index of $q$. Hence, $q^\prime = (p-2)/(p-4)$, implying that
$$
\E |h|^{2\alpha} |f|^{2(2+\alpha)} \leq (\E |h|^{2\alpha q^\prime})^{1/q^\prime} \cdot (\E |f|^{2(2+\alpha)q})^{1/q} = \|h\|_{L_2}^{2(1-2/(p-2))} (2M)^{2p/(p-2)};
$$
Hence,
$$
(\E |\xi h|^2)^{1/2} \leq c\|h\|_{L_2}^{1-\frac{2}{p-2}} \cdot M^{\frac{p}{p-2}},
$$
as claimed.
\endproof

Next, recall that the parameters in $(\heartsuit)$ and $(\spadesuit)$ are $\nu$---as in Remark \ref{rem:value-of-nu}, and $\gamma$. We show in what follows that in fact, one may take $\gamma=\nu$.

Recall that for a triplet $(U,X,Y)$
$$
\M_{u,u^*}(j)=\frac{2}{m} \sum_{i \in I_j} \xi_i (u-u^*)(X_i)
$$
where $\xi_i = (u^*(X_i)-Y_i)$.

\begin{Theorem} \label{thm:P3}
There exist absolute constants $c_0$, $c_1$ and $c_2$ for which the following holds.  Let $r$ satisfy that
\begin{equation} \label{eq:in-thm-P3}
\E \sup_{u \in U_{u^*,r}} \left|\frac{1}{N} \sum_{i=1}^N \eps_i \xi_i u(X_i) \right| \leq c_0r^2.
\end{equation}
Then with probability at least $1-2\exp(-c_1  n)$, for every $u \in U$
\begin{description}
\item{$\bullet$} if $\|u-u^*\|_{L_2} \geq r$ then
$$
|\M_{u,u^*}(j)-\E \M_{u,u^*} | \leq \nu \|u-u^*\|_{L_2}^2 \ \ {\rm for \ at \ least \ } 0.99n \ {\rm indices \ } j;
$$
\item{$\bullet$} if $\|u-u^*\|_{L_2} \leq r$ then
$$
|\M_{u,u^*}(j)-\E \M_{u,u^*} | \leq \nu r^2 \ \ {\rm for \ at \ least \ } 0.99n \ {\rm indices \ } j.
$$
\end{description}
\end{Theorem}

\proof
For $1 \leq j \leq n$ and $v \in L_2(\mu)$, let
$$
W_v(j) = \frac{2}{m} \sum_{i \in I_j} \left(v(X_i) \xi_i - \E v(X) \xi \right),
$$
which are all distributed as $W_v$.

It is enough to show that with high probability,
\begin{equation} \label{eq:in-proof-multi-Z}
Z=\frac{1}{n}\sup_{v \in U_{u^*,r}} |\{ j : |W_v(j)| \geq \nu r^2\} | \leq 0.01.
\end{equation}
Indeed, the wanted estimate for $\|v\|_{L_2} \geq r$ follows by considering
$$
v \in {\rm star}(U-u^*,0) \cap r S(L_2) \subset U_{u^*,r}
$$
and then invoking the star-shape property, combined with the fact that $W_v$ is homogenous in $v$. The estimate for $\|v\|_{L_2} < r$ is immediate from \eqref{eq:in-proof-multi-Z}.

\vskip0.3cm

Observe that by the bounded differences inequality (see, e.g., \cite{BoLuMa13}), the random variable $Z$ concentrates well around its mean: with probability at least $1-2\exp(-c_0n)$, $Z \leq \E Z + 0.001$. Next, note that
\begin{align*}
Z \leq & \frac{1}{\nu r^2} \sup_{v \in U_{u^*,r}}
\frac{1}{n}\sum_{j =1}^n |W_v(j)|
\\
\leq & \frac{1}{\nu r^2} \sup_{v \in U_{u^*,r}} \left| \frac{1}{n}\sum_{j=1}^n  |W_v(j)| - \E |W_v(j)| \right| + \frac{1}{\nu r^2} \sup_{v \in U_{u^*,r}} \E |W_v|,
\end{align*}
and by a symmetrization and contraction argument,
\begin{equation*}
\E Z \leq \frac{2}{\nu r^2} \E \sup_{v \in U_{u^*,r}} \left| \frac{1}{n} \sum_{j=1}^n  \eps_i W_v(j) \right| +  \frac{1}{\nu r^2} \sup_{v \in U_{u^*,r}} \E |W_v|.
\end{equation*}

Using Lemma \ref{lemma:L-p-multiplier} one has that
$$
\E |W_v| \leq \frac{2}{m} \E \left(\sum_{i \in I_j} v^2(X_i) \cdot \xi_i^2\right)^{1/2} \leq \frac{2\|\xi v\|_{L_2}}{\sqrt{m}} \leq c_1 \frac{M^{\frac{p}{p-2}} r^{1-\frac{2}{p-2}}}{\sqrt{m}}.
$$
Therefore,
$$
\frac{1}{\nu r^2} \sup_{v \in U_{u^*,r}} \E |W_v| \leq \frac{1}{\nu \sqrt{m}} \cdot \left(\frac{M}{r}\right)^{\frac{p}{p-2}} \leq 0.001,
$$
recalling that $\nu$ is an absolute constant and by the choice of $m$.

Finally, using that each $W_v$ is mean-zero, a symmetrization argument shows that
\begin{equation*}
\E \sup_{v \in U_{u^*,r}} \left| \frac{1}{n}\sum_{j=1}^n  \eps_j W_v(j) \right|  \leq  4\E \sup_{v \in U_{u^*,r}} \left| \frac{1}{N} \sum_{i=1}^N \eps_i \xi_i v(X_i)\right|,
\end{equation*}
implying that
$$
\E Z \leq \frac{8}{\nu r^2} \E \sup_{u \in F_{f^*,r}} \left| \frac{1}{N}\sum_{i=1}^N \eps_i \xi_i u(X_i) \right| \leq 0.001.
$$
Indeed, the last inequality holds provided that
\begin{equation} \label{eq:inproof-multi-cond1}
\E \sup_{v \in U_{u^*,r}} \left| \frac{1}{N}\sum_{i=1}^N \eps_i \xi_i v(X_i) \right| \leq  c r^2
\end{equation}
for a well-chosen constant $c$ that depends only on $\nu$---which is an absolute constant; therefore, $c$ is an absolute constant as well.
\endproof

The proof of Theorem \ref{thm:P3} completes the proof of Theorem \ref{thm:main}. To see that, let $(F,X,Y)$ be a triplet such that $F \subset M B(L_p(\mu))$ and $\|Y\|_{L_p} \leq M$ for some $p>4$. Consider any triplet $(U,X,Y)$ such that $U \subset \bar{F}$, and $f^* \in U$, and let $N \geq N_0$ for $N_0$ is as in Theorem \ref{thm:main}. Then $(\heartsuit)$ and $(\spadesuit)$ hold with $\nu$ and $\gamma$ that are absolute constants; $r=\sqrt{\eps}$ satisfies \eqref{eq:in-thm-P3}; and $1-2\exp(-c_1n) \geq 1-\delta/4$ as required.

\vskip0.3cm
Combining all the components it follows that any such triplet $(F,X,Y)$ is \emph{suitable} in the sense of Definition \ref{def:suitable}. Hence, by Theorem \ref{thm:prop-to-error}, $\E {\cal L}_{\hat{f}} \leq c\eps$ with probability at least $1-\delta$.
\endproof

\begin{Remark} \label{rem:p-2-4}
Lemma \ref{lemma:L-p-multiplier} is the only place in the proof in which the assumption that $p>4$ is actually used. In all the other components it suffices that $p>2$ (while keeping track of the way the constants depend on $p$; they become unbounded as $p$ approaches $2$). To make the proof work in the range $p \in (2,4]$ one has to ensure that if $\|u\|_{L_p}, \ \|\xi\|_{L_p} \leq M$ then
$$
\E \left(\frac{1}{m}\sum_{i=1}^m u^2(X_i) \xi_i^2\right)^{\frac{1}{2}} \leq C \phi(M,\|u\|_{L_2})
$$
for some function $\phi$ (in the case $p>4$ one may use $\phi(M,r) = c M^{p/(p-2)} r^{1-2/(p-2)}$). The identity of $\phi$ affects the choice of the $m$ and therefore, the required sample complexity estimate. Of course, that is merely a sufficient condition that makes this proof work. It is possible that the outcome of Theorem \ref{thm:main} is still true even if this proof fails.

We leave the study of what happens in the range $p \in (2,4]$ to future work.
\end{Remark}

\bibliographystyle{plain}
\bibliography{tournament3}

\end{document}